%% file: large_scale_init_main.tex
\documentclass[twoside]{article}

\usepackage{aistats2020}
\usepackage{natbib}

\usepackage[utf8]{inputenc} 
\usepackage[T1]{fontenc}    
\usepackage{hyperref}       
\usepackage{url}            
\usepackage{booktabs}       
\usepackage{amsfonts}       
\usepackage{nicefrac}       
\usepackage{microtype}      
\usepackage{amsmath, amsfonts, amssymb}
\usepackage{color}
\usepackage{subfigure}
\usepackage{graphicx}
\usepackage{adjustbox}

\usepackage{soul}

\usepackage{caption}

\begin{document}

%

%

\twocolumn[

\aistatstitle{If dropout limits trainable depth, does critical initialisation still matter? A large-scale statistical analysis on ReLU networks}

\aistatsauthor{ Arnu Pretorius$^{1}$ \And Elan van Biljon$^{1}$ \And  Benjamin van Niekerk$^{2}$ }
\aistatsauthor{ Ryan Eloff$^{2}$ \And Matthew Reynard$^{2}$ \And  Steven James$^3$ }
\aistatsauthor{ Benjamin Rosman$^{3,4}$ \And Herman Kamper$^{2}$ \And  Steve Kroon$^{1}$ }

\aistatsaddress{ $^1$Computer Science Division, Stellenbosch University \\ 
$^2$Department of E\&E Engineering, Stellenbosch University \\ 
$^3$Department of Computer Science and Applied Mathematics, University of the Witwatersrand \\
$^4$Council for Scientific and Industrial Research, South Africa} 

]

\input{abstract}
\input{introduction}
\input{background}

\input{expDesign}
\input{results}

\input{conclusion}

\clearpage
\newpage

\bibliographystyle{IEEEtranN}
\bibliography{bibfile}

\clearpage
\newpage

\input{suppmat.tex}


\end{document}

%% file: abstract.tex
\begin{abstract}

    Recent work in signal propagation theory has shown that dropout limits the depth to which information can propagate through a neural network.
    In this paper, we investigate the effect of initialisation on training speed and generalisation for ReLU networks within this depth limit.
    We ask the following research question: given that critical initialisation is crucial for training at large depth, if dropout limits the depth at which networks are trainable, does initialising critically still matter?
    We conduct a large-scale controlled experiment, and perform a statistical analysis of over $12000$ trained networks. 
    We find that (1) trainable networks show no statistically significant difference in performance over a wide range of non-critical initialisations; (2) for initialisations that show a statistically significant difference, the net effect on performance is small; (3) only extreme initialisations (very small or very large) perform worse than criticality. 
    These findings also apply to standard ReLU networks of moderate depth as a special case of zero dropout. 
    Our results therefore suggest that, in the shallow-to-moderate depth setting, critical initialisation provides zero performance gains when compared to off-critical initialisations and that searching for off-critical initialisations that might improve training speed or generalisation, is likely to be a fruitless endeavour. 
\end{abstract}

%% file: introduction.tex
\section{Introduction}

Dropout is arguably one of the most popular and successful forms of regularisation for deep neural networks~\citep{srivastava2014dropout}. 
This has sparked research into analysing dropout's effects \citep{wang2013fast, wager2013dropout, baldi2013understanding}, extending dropout's mechanism of regularisation \citep{wan2013regularization, gal2017concrete, gomez2018targeted, ghiasi2018dropblock} and connecting dropout to different Bayesian inference methods \citep{kingma2015variational, gal2016dropout, molchanov2017variational}. 
Despite its success, dropout has also been shown to limit the \textit{trainable depth} of a neural network \citep{schoenholz2016deep}.

At initialisation, the random weight projection at each layer combined with dropout may cause inputs to become uniformly correlated beyond a certain depth. 
Thus discriminatory information in the inputs may vanish before reaching the output layer. 
The trainable depth of a network is the maximum depth to which this information is able to propagate forward without completely vanishing in this way. 
\cite{schoenholz2016deep} arrive at this result through a \textit{mean field} analysis of dropout at initialisation. 

Mean field theory provides a powerful approach to analysing deep neural networks and has become a cornerstone of recent discoveries in improved initialisation schemes.
These schemes, often referred to as \textit{critical initialisations}, ensure stable signal propagation dynamics by preserving second moment input statistics during the forward pass, even at infinite depth.
Critical initialisation has made it possible to train \textit{extremely deep} networks (sometimes up to 10000 layers) for a variety of different architectures \citep{pennington2017resurrecting, xiao2018dynamical, chen2018dynamical}. 
Using the tools of mean field theory, \citet{pretorius2018critical} extend these results to fully-connected ReLU networks with multiplicative noise regularisation.
These results hold for a general class of noise distributions, while earlier work~\citep{hendrycks2016adjusting} describes dropout-specific initialisation schemes. 

For non-critical initialisation, signal propagation can become unstable and result in the saturation of activation functions.
In the particular case of ReLU activations, numerical instability (overflow or underflow) can arise when training very deep networks.
Despite this, when training ReLU networks of a finite depth, there is a range of trainable but non-critical initialisations.
That is, there exists a ``band'' of valid initialisations around the critical point.
It is conceivable that using these alternative, non-critical initialisations may confer some benefits. 
For example, \citet{saxe2013exact} note that just off of criticality, the spectrum of the input-output Jacobian can be well behaved, which has been linked to improvements in training and generalisation \citep{pennington2017resurrecting}.
This leads us to the following question.  

\textbf{Question}: 
\textit{If dropout limits the depth to which networks can train, does critical initialisation still matter?} 
Given that stable signal propagation at extreme depths is no longer a concern, are there alternative initialisations that might perform better than the critical initialisation? 



To investigate the above research question, we conduct a large-scale randomised control trial (RCT)---an approach borrowed from the medical community---to compare training speed and generalisation for ReLU neural networks with dropout for different initialisations. 
We consider multiple datasets, training algorithms, dropout rates and combinations of 
hyperparameters to avoid confounding effects.
To the best of our knowledge, this is the first application of RCTs in a deep learning context.
A statistical analysis of our results leads to the following insight.

\textbf{Answer}: \textit{There is no statistically significant difference between the critical initialisation and a wide neighbourhood of non-critical initialisations, as measured by training speed and generalisation.} 
In our experiment we find that this also applies to standard ReLU networks without dropout, for which the critical initialisation is the popular ``He'' initialisation \citep{he2015delving}.
Our findings seem to indicate that networks of moderate depth (less than 20 layers) are in fact very insensitive to initialisation. 
In addition, we conclude that exploring the initialisation landscape around criticality in the hope of finding previously undiscovered benefits, is unlikely to be a fruitful enterprise. 

%% file: background.tex
\section{Background}


We model the expected value of a target variable $\mathbf{y}$ conditioned on an input $\mathbf{x}$, i.e. $\mathbb{E}(\mathbf{y}| \mathbf{x})$, using a fully-connected feedforward neural network with dropout. 
Given an input $\mathbf{x}^0 \in \mathbb{R}^{D_0}$, we can define this neural network recursively as 
\begin{align}
	\mathbf{x}^l = \phi(\tilde{\mathbf{h}}^l), \textcolor{white}{space} \tilde{\mathbf{h}}^l = W^l\left(\mathbf{x}^{l-1}\odot \frac{\epsilon^{l-1}}{1-\theta}\right) + \mathbf{b}^l,
	\label{eq: noisy deep model}
\end{align}
for $l = 1, ..., L$, where $L$ is the total number of hidden layers, $\odot$ denotes element-wise multiplication, and $\epsilon^l \sim \textrm{Bern}(1-\theta)$ is a Bernoulli noise vector, corresponding to a dropout rate $\theta$. 
The dimensionality of hidden layer $l$ is denoted as $D_l$, and activations at each layer are computed element-wise using $\phi(a) = \textrm{ReLU}(a) = \max(0, a)$. 
The initial weights $W^l \in \mathbb{R}^{D_l \times D_{l-1}}$ and biases $\mathbf{b}^l \in \mathbb{R}^{D_l}$ are sampled i.i.d.\ from zero-mean Gaussian distributions with variances $\sigma^2_w/D_{l-1}$ and $\sigma^2_b$, respectively. 

We focus on ReLU because of its widespread use and empirical success and consider the fully-connected setting since derived conclusions for these networks often generalise to other architectures, e.g. convolutional networks \citep{he2015delving, xiao2018dynamical}. 
Through their work, \cite{schoenholz2016deep} also hypothesise that the signal propagation behaviour of many different architectures is likely to be governed by the fully-connected case.
 
\begin{figure*}
	\centering
	\includegraphics[width=0.7\linewidth]{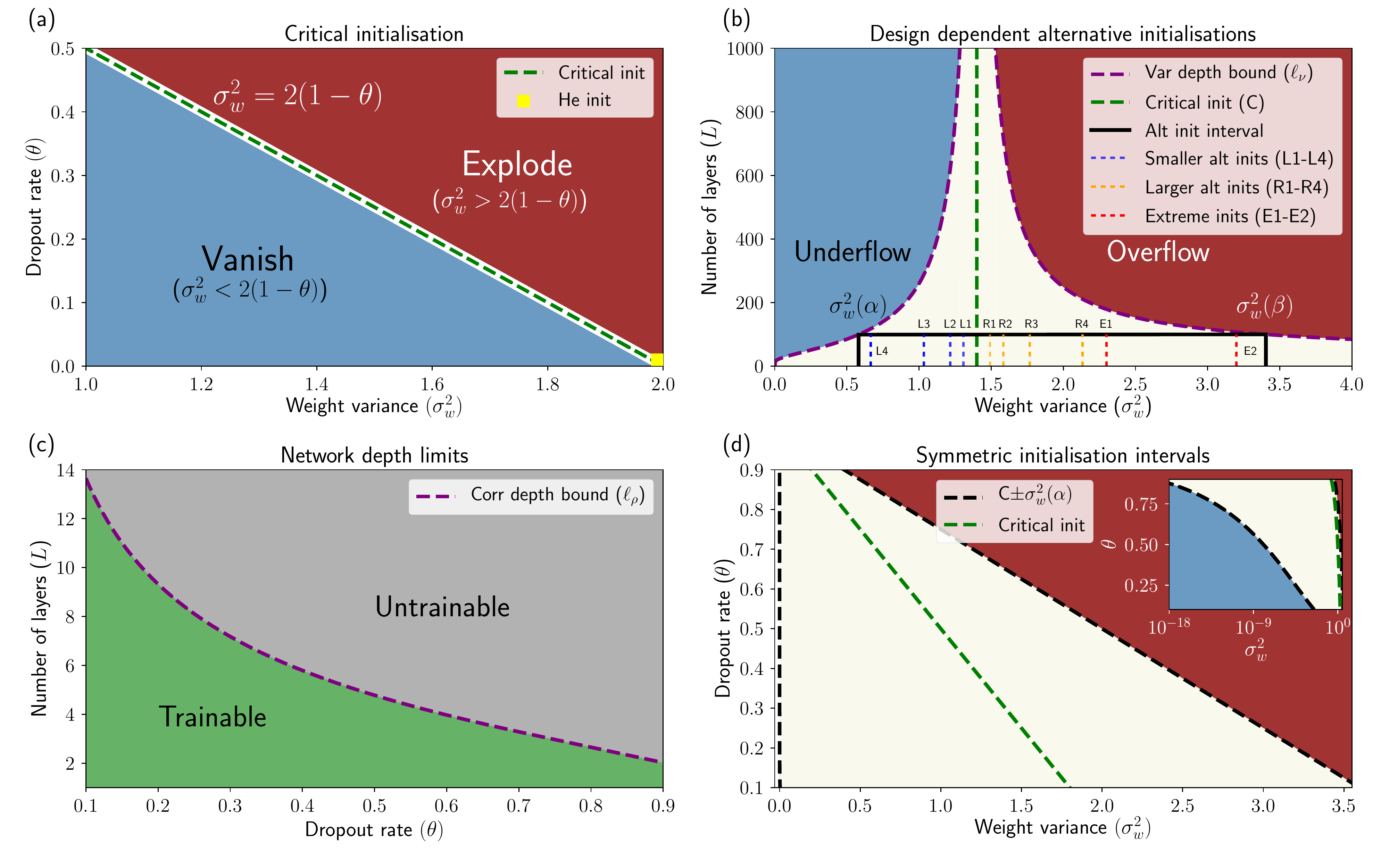}
	\caption{\textit{Choosing network depth and initialisation}. \textbf{(a)}: Critical initialisation boundary separating regimes of vanishing or exploding variance signal propagation at large depth. \textbf{(b)}: An illustration of the region that allows stable variance (beige) and correlation (black bordered beige) information to propagate through the entire network. (We choose a depth $L=100$ for ease of visualisation, although this depth is too deep for training with dropout). Alternative non-critical initialisations L1--L4 (blue), C (green), R1--R4 (orange), E1--E2 (red) are sampled from this region. \textbf{(c)}: The depth to which networks with dropout are trainable for different dropout rates (dashed purple line). \textbf{(d)} Symmetrical interval containing L1--L4 and R1--R4 as a function of the dropout rate: the (beige) region around criticality represents the set of trainable initialisations.}
	\label{fig: theory}
\end{figure*}

\subsection{Mean field theory for signal propagation} 

\citet{poole2016exponential}, \citet{schoenholz2016deep} and \citet{pretorius2018critical} use mean field theory to analyse fully-connected feedforward neural networks at initialisation. 
For large layer widths, each pre-activation (the linear combination of the incoming connections from the previous layer) at initialisation in any given layer of the network represents a large sum of i.i.d.\ random variables.
According to the central limit theorem, this sum will tend to a Gaussian distribution in the limit of infinite width. 
Using the above observation, the mean field approach is to fit Gaussian distributions over all the pre-activation units through moment matching to describe the behaviour of wide random neural networks at initialisation. 

In more detail, consider two inputs $\mathbf{x}^0_1$ and $\mathbf{x}^0_2$. 
Denote the scalar pre-activation at unit $j$ in layer $l$ for input $\mathbf{x}^0_1$ as $\tilde{h}^{l,1}_j$. 
For fully-connected ReLU networks with dropout, \cite{pretorius2018critical} derive the joint distribution over the pre-activations in expectation over the network parameters and the noise as 
\begin{align*}
	p\left (\tilde{h}^{l, 1}_j, \tilde{h}^{l, 2}_{j} \right ) = \mathcal{N}(\mathbf{0}, \tilde{\Phi}^l),
\end{align*}
where
\begin{align*}
	\tilde{\Phi}^l & = \begin{bmatrix}
		\nu^l_{1} & \kappa^l \\
		\kappa^l & \nu^l_{2}
	\end{bmatrix}.
\end{align*}
The layer-wise evolution of the terms in the covariance matrix $\tilde{\Phi}^l$, are given by
\begin{align}
\nu^l_1 & = \frac{\sigma^2_w}{2(1-\theta)}\nu_1^{l-1}  + \sigma^2_b \label{eq: variance recurrence}\\ 
\kappa^l & = \frac{\sigma^2_w}{2}\kappa^{l-1}\left(g(\rho^{l-1}) + \frac{1}{2}\right) + \sigma^2_b \\ 
\rho^l & = \kappa^l / \sqrt{\nu^l_1 \nu^l_2} \label{eq: correlation recurrence} 
\end{align}
where
$$g(\rho^{l-1}) = \frac{1}{\pi\rho^{l-1}} \left ( \rho^{l-1}\sin^{-1} \left (\rho^{l-1} \right ) + \sqrt{1 - (\rho^{l-1})^2} \right)$$
with initial variance $\nu^0_1 = \frac{\mathbf{x}^0_1\cdot\mathbf{x}^0_1}{D^0}$ and covariance $\kappa^0 = \frac{\mathbf{x}^0_1\cdot\mathbf{x}^0_2}{D^0}$. 
The  above quantities are derived in the large width limit, but in practice  tend to hold for finite widths of moderate size \citep{poole2016exponential, schoenholz2016deep, pretorius2018critical}. 

\paragraph{Critical initialisation.} A fixed point of the variance recurrence in \eqref{eq: variance recurrence} is given by
\begin{align*}
\{\sigma^2_w, \sigma^2_b\} = \{2(1-\theta), 0\},
\end{align*}
which ensures that signal propagation variances are preserved during the forward pass of a ReLU network with dropout \citep{pretorius2018critical}. 
These settings of the network parameters is referred to as the \textit{critical initialisation}. Figure \ref{fig: theory}(a) shows the relationship between the critical initialisation and the dropout rate. 
Away from criticality, the variance signal tends to vanish or explode. 
If the dropout rate $\theta$ is zero, the initialisation reduces to the popular ``He'' initialisation for ReLU networks \citep{he2015delving}.

\paragraph{Trainable depth.} Consider the following proposition due to \citet{schoenholz2016deep}:
\begin{quote}
	\textbf{Proposition 1}: \textit{At initialisation, a necessary condition for training any neural network is that the information from the input layer should be able to reach the output layer}. 
\end{quote}
\citet{pretorius2018critical} analyse the evolution of the input variances and correlations, as given in \eqref{eq: variance recurrence} and \eqref{eq: correlation recurrence}, to establish when this information propagation requirement is violated.
Specifically, let $\alpha$ and $\beta$ represent the smallest and largest positive values representable on a modern machine. 
The depth at which numerical instability issues (underflow or overflow) arise from the variances described in \eqref{eq: variance recurrence} for non-critical initialisations is bounded by
\begin{align}
	\ell_{\nu} = \begin{cases}
		\frac{\ln\left(\frac{\alpha}{\nu^0}\right)}{\ln\left(\frac{\sigma^2_w}{2(1-\theta)}\right)}, \textrm{ if } \sigma^2_w < 2(1-\theta) \\
		\frac{\ln\left(\frac{\beta}{\nu^0}\right)}{\ln\left(\frac{\sigma^2_w}{2(1-\theta)}\right)}, \textrm{ if } \sigma^2_w > 2(1-\theta). 
		\label{eq: variance bounds}
	\end{cases}
\end{align} 
An example of these bounds are shown in Figure \ref{fig: theory}(b) as purple dashed lines. 
The depth bounds around criticality are finite but large, exceeding typical depths for most modern deep neural networks used in practice. 
However, even if single input information can propagate to large depths, the correlation between inputs, described in \eqref{eq: correlation recurrence}, converge to degenerate levels over a much shorter depth horizon \citep{pretorius2018critical}. 
This can limit the network's ability to train, since all discriminatory information is lost during forward propagation.
Furthermore, the rate of convergence in correlation is invariant to initialisation, but increases as more dropout is applied.  
As a result, inputs to a dropout network tend to convey similar information at shallower depths compared to unregularised networks. 
The bound that characterises convergence in correlation is 
\begin{align}
	\ell_\rho = -6/\ln \left[\frac{(1-\theta)}{\pi}\left ( \sin^{-1}(\rho^*) + \frac{\pi}{2} \right ) \right],
	\label{eq: depth scales}
\end{align}
where $\rho^*$ denotes the converged correlation and the factor $6$ is an ad-hoc factor, which seems to provide a good fit to experimental data, but is as yet unexplained \citep{schoenholz2016deep, pretorius2018critical}. 
Figure \ref{fig: theory}(c) plots the theoretically predicted trainable depth using \eqref{eq: depth scales} for different dropout rates. 
Note that these depths are much shallower than those derived for variance dynamics.

%% file: expDesign.tex
\section{Experimental setup}

We conduct a large-scale controlled experiment using networks of trainable depth to compare the effect of initialisation on training speed and generalisation for ReLU networks with dropout. 
We explore the space around criticality by selecting alternative initialisations whose values theoretically satisfy Proposition 1.
Our final aim is to test whether there exists a statistically significant difference, as measured by training speed and generalisation, between the different initialisations.
To answer this question, we use a systematic randomised control trial methodology with hypothesis testing.

\subsection{Controlled experiments using neural networks: a randomised control trial approach}

Inspired by causal discovery in medical research, we consider a hypothesis a priori and conduct a ``randomised control trial'' (RCT) \citep{kendall2003designing} using neural networks.
In an ordinary randomised control trial a random sample, representative of the full population, is split into two groups. One group receives some form of an intervention, such as a new drug.
The other group, referred to as the control group, receives no intervention. 
The purpose of the two groups is to control for all confounding effects that are unrelated to the intervention of interest. 
The groups are then monitored by collecting data over time. 
Once the study has been completed, a test for statistical significance can be applied to ascertain if there exists a difference between the two groups, as measured by a quantitative metric of interest. 
If a statistical significant difference is detected, the intervention is confirmed as being the cause. 
In this paper, we aim to test for differences in initialisation of fully-connected ReLU neural networks with dropout. 

\begin{figure*}
    \centering
    \includegraphics[width=0.7\linewidth]{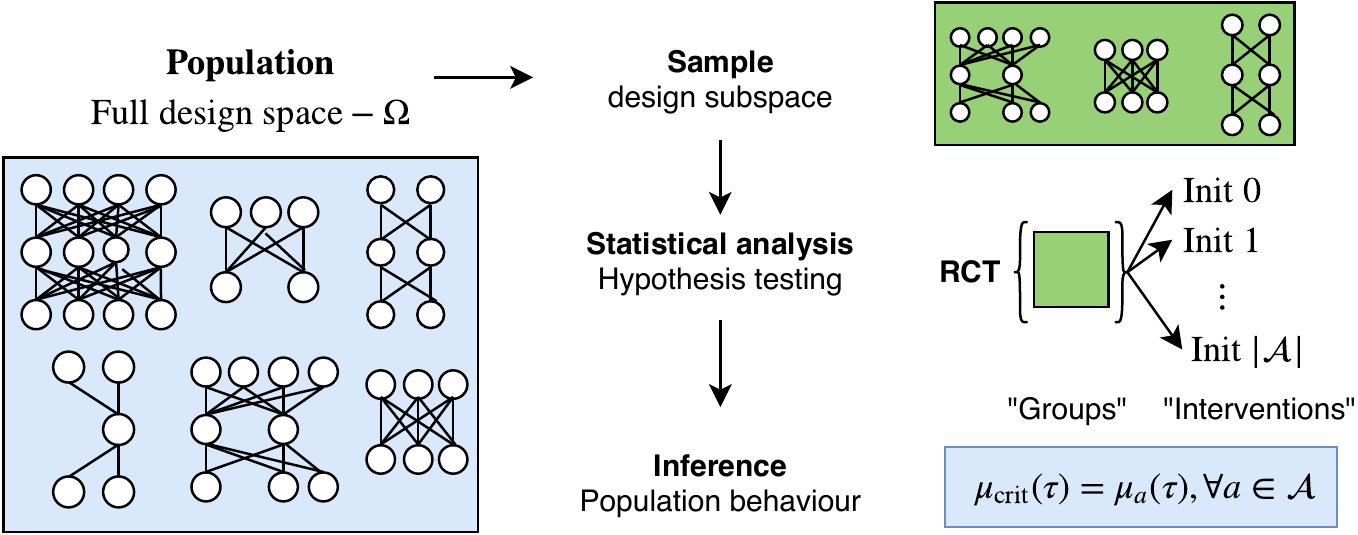}
    \caption{Randomised control trial approach to analysing the effect of initialisation in neural networks.}
    \label{fig: rct}
\end{figure*}


To begin, consider the following \textit{design space}: 
\begin{quote}
\textbf{$\Omega$-design space:} \textit{We define the neural network design space $\Omega$ as the space consisting of different possible combinations of design components used to construct an algorithm for classification using a fully-connected ReLU neural network with dropout. Specifically, the design space is given by the following Cartesian product
\begin{align*}
    \Omega = \mathcal{X} \times \mathcal{D} \times \mathcal{W} \times \mathcal{R} \times \mathcal{B} \times \mathcal{O} \times \mathcal{M} \times \mathcal{L}
\end{align*}
where the component sets divide into (1) dataset $\mathcal{X}$, (2) network topology: depth $\mathcal{D}$, width $\mathcal{W}$, (3)~dropout rate $\mathcal{R}$, and (4) training procedure: batch size~$\mathcal{B}$, optimiser $\mathcal{O}$, momentum $\mathcal{M}$, and learning rate $\mathcal{L}$.} 
\end{quote}

We adapt the RCT approach for analysing neural network initialisation as follows. 
First, we randomly generate a collection of different neural network algorithms by sampling from the design space, or ``population,'' of possible neural networks. 
For example, a 10-layer ReLU network trained on MNIST, where each layer is 256 units wide, with a dropout rate of $0.5$, optimised using RMSprop with zero momentum and a learning rate of $5\times10^{-4}$ and batches of size $128$, corresponds to the 8-tuple: $(\textrm{MNIST}, 10, 256, 0.5, 128, \textrm{RMSprop}, 0, 5\times10^{-4})$. 
Next, we construct identical ``groups'' by using multiple copies of the sampled designs.
Each group in the experiment is then assigned a different initialisation scheme.
Finally, we test the following hypothesis related to a given metric:
\begin{quote}
    \textbf{Null hypothesis}: \textit{Given a metric $\tau$, let $\mu_{\textrm{crit}}(\tau)$ denote the group mean associated with the critical initialisation and $\mu_{a}(\tau)$, the mean associated with an alternative initialisation $a \in \mathcal{A} \subset \mathcal{I}$, where $\mathcal{I}$ is the set of all possible initialisations, and $\mathcal{A}$ is our chosen set of alternative initialisations. Then the null hypothesis to be tested is} 
    \begin{align}
        \textrm{H}_0: \mu_{\textrm{crit}}(\tau) = \mu_a(\tau), \forall a \in \mathcal{A}. \label{eq: null hypotheses}
    \end{align}
\end{quote}
We discuss our methodology for selecting alternative initialisations in Section \ref{sec: alt inits}.

If the null hypothesis is rejected, we have strong evidence to indicate that the performance of the critical initialisation is significantly different from those of alternative initialisations. 
If $\textrm{H}_0$ cannot be rejected, the perceived difference is not considered statistically significant. 
Figure \ref{fig: rct} summarises this approach to studying the effect of initialisation in neural networks.

\paragraph{Metrics.} Our chosen metrics of interest are training speed and generalisation performance. Specifically, we define these quantities as
\begin{itemize}
    \item \textbf{Training speed -- $\tau_{s}$}: accuracy achieved on the training set at the $100\textsuperscript{th}$ epoch.
    \item \textbf{Generalisation -- $\tau_{g}$}: highest accuracy achieved on the test set over the course of training.
\end{itemize}
For example, $\mu_{\textrm{crit}}(\tau_{s})$ denotes the mean training speed associated with the critical initialisation, where a higher mean accuracy at epoch $100$ indicates faster training. 

\paragraph{Sampling algorithm designs.} For our experiment groups, we sample $1120$ different designs. These designs are drawn randomly from $\Omega$, which we construct by forming the Cartesian product of the following discrete sets for dataset, depth, width, dropout rate, batch size, optimiser, momentum and learning rate:
\begin{align*}
    \mathcal{X} & = \{\textrm{MNIST}, \textrm{FashionMNIST}, \textrm{CIFAR-10}, \textrm{CIFAR-100}\} \\ 
    \mathcal{D} & = \{2, 3, 4, 5, 6, 7, 8, 9, 10, 11, 12, 13, 15, 20\} \\ 
    \mathcal{W} & = \{400, 600, 800\} \\
    \mathcal{R} & = \{\textrm{0, 0.1, 0.3, 0.5}\}\\
    \mathcal{B} & = \{32, 64, 128, 256\} \\
    \mathcal{O} & = \{\textrm{SGD}, \textrm{Adam}, \textrm{RMSprop} \} \\ 
    \mathcal{M} & = \{0, 0.5, 0.9\} \\
    \mathcal{L} & = \{10^{-3}, 10^{-4}, 10^{-5}, 10^{-6}\}
\end{align*}
For a given dropout rate, we limit the sampled depths in $\mathcal{D}$ to only the settings that would allow useful correlation information to reach the output layer, i.e. $d \leq \ell_\rho, \forall d \in \mathcal{D}$. 
We also include network depths of $15$ and $20$ when no dropout is being applied. 
We sample $70$ designs for each dropout rate and dataset combination for a large enough diversity in network architecture and optimisation.
To ensure a balanced set of network designs, we simply duplicate each group of designs for every dropout rate in $\mathcal{R}$, as well as for each dataset. 
A full description of this process is presented in Appendix A.
Finally, each network is trained for $500$ epochs on MNIST \citep{lecun1998gradient}, FashionMNIST \citep{xiao2017fashion}, CIFAR-10 and CIFAR-100 \citep{krizhevsky2009learning}, using the full training set for each.


\subsection{A network design dependent set of alternative initialisations} \label{sec: alt inits}

We ensure that our networks preserve short range correlation information by limiting their depth. 
To develop a principled approach towards exploring the initialisation space around criticality, we now ask: for a fixed depth, what is the range of initialisations around criticality that will remain numerically stable until the output layer? 
In other words, for which $\sigma^2_w$ in \eqref{eq: variance bounds} is $\ell_\nu \geq \ell_\rho$. 
We can find these bounds for alternative initialisations by solving for $\sigma^2_w$ in  \eqref{eq: variance bounds}, which gives

\begin{align}
\sigma^2_w(\alpha) & = \inf \left \{ \sigma^2_w \in \mathbb{R}_{> 0} \bigg | \ell_\nu \geq \ell_\rho, \sigma^2_w < 2(1-\theta) \right \} \nonumber \\
& = 2(1-\theta) \left (\frac{\alpha}{\nu^0}\right )^{1 / \ell_\rho} \textrm{ [lower bound]}\label{eq: lower bound}\\
\sigma^2_w(\beta) & = \sup \left \{ \sigma^2_w \in \mathbb{R}_{> 0} \bigg | \ell_\nu \geq \ell_\rho, \sigma^2_w > 2(1-\theta)  \right \} \nonumber \\
& = 2(1-\theta) \left (\frac{\beta}{\nu^0}\right )^{1 / \ell_\rho} \textrm{ [upper bound]}\label{eq: upper bound}
\end{align}

In our experiments, we use $32$-bit floating point precision, such that $\alpha = 1.1754944 \times 10^{-38}$ in \eqref{eq: lower bound} and $\beta = 3.4028235 \times 10^{38}$ in \eqref{eq: upper bound}. 
An example interval for possible alternative initialisations bounded by $\sigma^2_w(\alpha)$ and $\sigma^2_w(\beta)$ is shown in Figure \ref{fig: theory}(b). Note that the interval is not symmetric. 
This is because although the signal vanishes and explodes at the same rate, the critical initialisation is typically much closer to $\alpha$ than to $\beta$. 
This causes the interval to be wider to the right. To cover this entire space around criticality would be computationally infeasible. 
Therefore, we focus on sampling alternative initialisations around criticality as a function of the dropout rate. 

Specifically, we first sample a core set of initialisations within the interval $\textrm{C} \pm \sigma^2_w(\alpha)$ centred around the critical initialisation (C), with logarithmic spacing between samples (see Appendices B and C for more detail). 
This symmetric interval is illustrated by the dashed black lines in Figure \ref{fig: theory}(d) for different dropout rates. 
Note that the interval becomes narrower for larger dropout rates.
The inset in Figure~\ref{fig: theory}(d) plots the left side of the interval close to zero on a log-scale.
The core set of alternative initialisations for the fixed depth in Figure~\ref{fig: theory}(b) are shown as blue dashed lines (below criticality, marked L1-L4) and orange dashed lines (above criticality, marked R1-R4).  
Finally, we explore further to the right by sampling halfway, as well as close to the end of the interval, between criticality and $\sigma^2_w(\beta)$.
These more extreme initialisations are depicted in red (marked E1 and E2).

\subsection{Statistical comparison methodology}

The null hypothesis $\textrm{H}_0$ 
can be tested using an \textit{omnibus test}, which is specifically designed for multiple comparisons \citep{demvsar2006statistical}. 
If the null hypothesis is rejected in this setting, there is evidence to suggest that at least one of the competing initialisations is significantly different from the rest. 
Specifically, we use the Iman-Davenport extension \citep{iman1980approximations} of the non-parametric Friedman rank test \citep{friedman1937use} as recommended by \citet{demvsar2006statistical} and \citet{garcia2010advanced}. 
We describe this test below in the context of comparing different initialisations.

\begin{itemize}
    \item \textbf{Friedman} \citep{friedman1937use}: For a given metric $\tau$ and a set of competing initialisations $\mathcal{I}$, the Friedman test first ranks initialisations $i \in \mathcal{I}$ in terms of their mean performances $\mu_i(\tau)$ and then computes a test statistic using these ranks. An average rank is assigned to tied initialisations. In more detail, let $r_{di}$ denote the rank for a specific design $d$ (sampled from $\Omega$) using initialisation $i$. We denote the mean rank over the set of all sampled designs $\Delta \subset \Omega$, as $\bar{R}_i = \frac{1}{|\Delta|} \sum^{|\Delta|}_{d=1}r_{di}$. The Friedman test statistic under the null hypothesis of no difference is
    \begin{align*}
        \chi^2_F = \frac{12|\Delta|}{|\mathcal{I}|(|\mathcal{I}| + 1)}\left ( \sum^{|\mathcal{I}|}_{i=1}\bar{R}^2_i - \frac{|\mathcal{I}|(|\mathcal{I}| + 1)^2}{4} \right ),
    \end{align*}
    and is approximately $\chi^2$ distributed with $|\mathcal{I}|-1$ degrees of freedom.
    \item \textbf{Iman-Davenport} \citep{iman1980approximations}: It has been shown that the Friedman test can be a poor approximation to the $\chi^2$ distribution. Therefore, the Iman-Davenport test modifies the Friedman test as follows
    \begin{align*}
        F_{ID} = \frac{(|\Delta| - 1) \chi^2_F}{|\Delta|(|\mathcal{I}|-1) - \chi^2_F},
    \end{align*}
    to more accurately approximate an $F$ distribution with $(|\mathcal{I}| - 1)$ and $(|\mathcal{I}|-1)(|\Delta|-1)$ degrees of freedom.
\end{itemize}

If we reject $\textrm{H}_0$, we may next ask whether there exists specific differences between the critical and the alternative initialisations. 
For this purpose we perform multiple pairwise tests. 

It is important to note, however, that when conducting multiple pairwise comparisons with popular two-sample tests, a significant difference might be detected simply by chance. 
To illustrate this, consider the probability of rejecting the null hypothesis when it is in fact true. 
This is known as a type I error. 
The null hypothesis is usually rejected if the probability of a type I error---the $p$-value---is less than some specified \textit{significance level}, typically set at $5\%$. 
However, it is insufficient to separately control for type I errors for each individual pairwise comparison. 
In our case, pairwise comparisons between the critical and the alternative initialisations (L1--L4, R1--R4, E1, E2) result in a total of $10$ comparisons.
At a significance level of $5\%$, a satisfactory probability of not making a type I error in a \textit{single} comparison is $\gamma = 1 - p(\textrm{reject H}_0 | \textrm{H}_0 \textrm{ is true}) = 95\%$. 
However, the probability of not making a type I error \textit{across all comparisons} is actually $\gamma^{10} \approx 60\%$, which is much lower than what was previously considered acceptable. 
Therefore, we guard against type I errors in multiple tests by using \textit{post-hoc} tests that aim to adjust the significance level to control the \textit{family-wise} error---the probability that at least one type I error is made among multiple tests \citep{garcia2010advanced, santafe2015dealing}. The specific post-hoc test we use is the Finner test \citep{finner1993monotonicity} as recommended by \citet{garcia2008extension} and \citet{garcia2010advanced}. 
The specifics of this test are given below.
\begin{itemize}
    \item \textbf{Finner} \citep{finner1993monotonicity}: Let $p_i$, $i=1, ..., i^*, ..., |\mathcal{I}|$, denote ordered $p$-values obtained from multiple pairwise comparisons corresponding to the null hypotheses of no mean difference $\textrm{H}_{01}, ..., \textrm{H}_{0i^*}, ..., \textrm{H}_{0|\mathcal{I}|}$. Using the Finner test, we reject $\textrm{H}_{01}, ..., \textrm{H}_{0i^*}$, where
    \begin{align*}
        i^* = \min \left \{i | p_i > 1 - \gamma^{i/(|\mathcal{I}| - 1)} \right \}.
    \end{align*}
\end{itemize}

\subsection{Summary of experimental setup} \label{sec: exp design summary}

We aim to test for differences in initialisation by conducting a large scale randomised control trial experiment using neural networks. 
We begin by sampling $70$ neural network algorithm designs from the design space $\Omega$ for each dropout rate and dataset combination, for a total of $1120$ designs.
To form groups in our experiment, we make $11$ identical copies of the $1120$ designs. 
Note that this is a core aspect of our approach.
We ensure \textit{within-group variation} by sampling \textit{different designs}, but then duplicate this collection of designs to form identical groups, one for each ``intervention'', i.e. initialisation.
For each group, we assign a different initialisation---either critical initialisation (C), or one of the $10$ alternative initialisations (L1--L4, R1--R4, E1--E2). 
All designs are then trained, resulting in a total of $70 \times 4 \times 4 \times 11 = 12320$ trained neural networks. 
Using these results, we test our hypothesis---that no difference exists between the various initialisations in terms of training speed and generalisation---using omnibus and post-hoc statistical tests.

To provide an analogy in the context of drug testing: our approach is akin to selecting a large random sample of human participants, duplicating (cloning) them to form identical groups, and then administering a different drug to each group.
To have exact copies of a representative sample to test on is an ideal case for an experimenter, since (1) within-group variation controls for confounding effects, and (2)
having identical groups ensures that if differences between groups are detected, it can only be as a result of the drug.
Here we are fortunate to be dealing with software entities and not human beings, which allows us the luxury of this ideal setup (see Appendix E for a further discussion on the validity of the RCT approach).

%% file: results.tex
\section{Results}

\begin{figure*}
\centering
\begin{subfigure}
  \centering
  \includegraphics[width=0.4\linewidth]{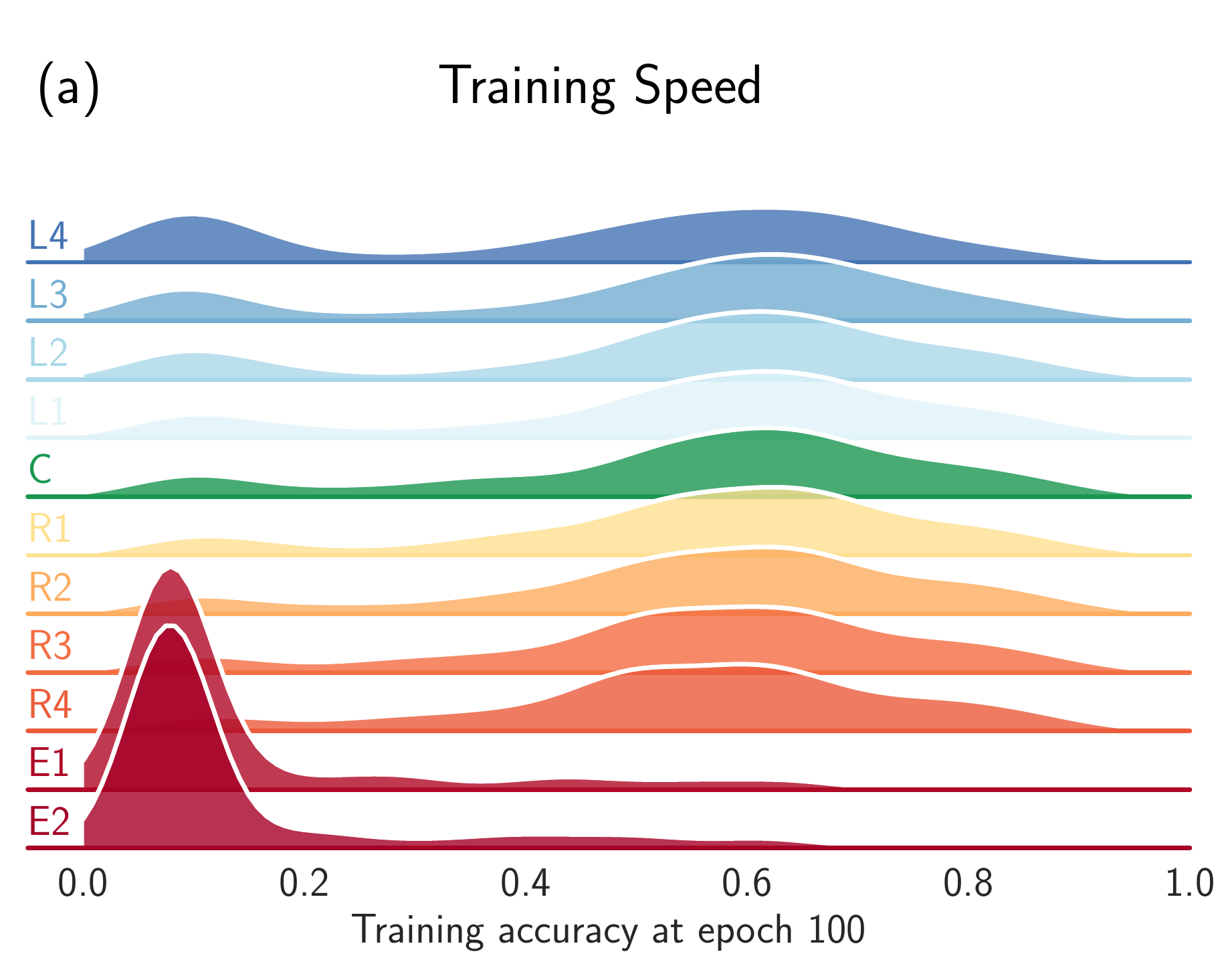}
  \label{fig:sub1}
\end{subfigure}%
~
\begin{subfigure}
  \centering
  \includegraphics[width=0.4\linewidth]{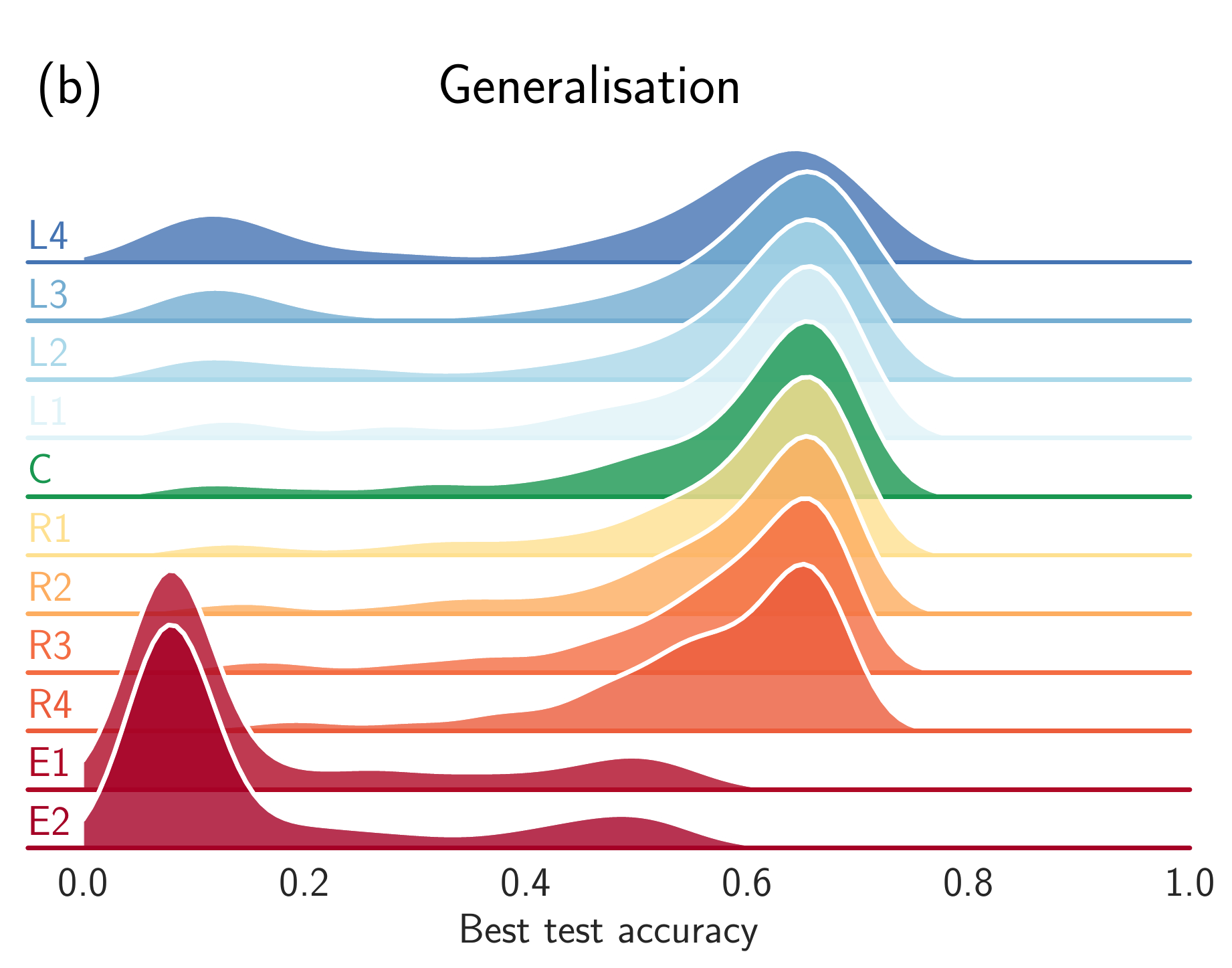}
  \label{fig:sub2}
\end{subfigure}
	\caption{\textit{Density fits for the different initialisations}. \textbf{(a)}: training speed. \textbf{(b)}: generalisation. }
	\label{fig: results}
\end{figure*}

A visualisation of our findings is presented in Figure \ref{fig: results}. 
For each initialisation, we plot densities summarising the results for (a)~training speed and (b)~generalisation. 
Visually our analysis seems to indicate that the average effect on training speed and generalisation for the critical initialisation is quite similar to the average effect of alternative initialisations, except at the extremes. 
To make this conclusion more concrete, we conduct a statistical analysis of the results.

\begin{table*}
	\caption{Post-hoc tests for training speed and generalisation. We use the symbols $^*$ to indicate a significant $p$-value (less than~$5\%$) and $^\dagger$ for a large effect size.}
	\label{tab: h tests}
	\begin{center}
		\begin{small}
			\begin{sc}
				Training speed -- $\textrm{H}_0: \mu_{\textrm{crit}}(\tau_s) = \mu_a(\tau_s), a \in \{L_i, R_i, E_1, E_2\}^4_{i=1}$
				\vspace{0.2cm}
			    \begin{adjustbox}{width=\linewidth,center}
				    \begin{tabular}{ccccccccccc}
					    \toprule
					    & L4 & L3 & L2 & L1 & R1 & R2 & R3 & R4 & E1 & E2 \\
						\toprule
						Adjusted $p$-value & $^*\mathbf{0}$ & $0.1338$  & $0.857$ & $0.4003$ & $0.6479$ & $0.0677$ & $^*\mathbf{3.95 \times 10^{-6}}$ & $^*\mathbf{3.33 \times 10^{-15}}$ & $^*\mathbf{2.4 \times 10^{-9}}$ & $^*\mathbf{2.4 \times 10^{-9}}$ \\
						Effect size & $-0.2512$ & $-0.0648$ & $-0.0287$ & $-0.0069$ & $0.0094$ & $0.0202$ & $0.0147$ & $-0.0048$ & $^\dagger\mathbf{-1.1055}$ & $^\dagger\mathbf{-1.1019}$ \\
					    \bottomrule
					\end{tabular}
				\end{adjustbox}
				
				\begin{sc}
					Generalisation -- $\textrm{H}_0: \mu_{\textrm{crit}}(\tau_g) = \mu_a(\tau_g), a \in \{L_i, R_i, E_1, E_2\}^4_{i=1}$ \\
						\begin{adjustbox}{width=\linewidth,center}
							\begin{tabular}{cccccccccccc}
								\toprule
								& L4 & L3 & L2 & L1 & R1 & R2 & R3 & R4 & E1 & E2 \\
								\toprule
								Adjusted $p$-value & $^*\mathbf{0}$ & $0.0545$ & $^*\mathbf{0.0033}$ & $^*\mathbf{0.0418}$ & $0.1226$ & $^*\mathbf{9.23 \times 10^{-6}}$ & $^*\mathbf{4.44 \times 10^{-16}}$ & $^*\mathbf{0}$ & $^*\mathbf{0}$ & $^*\mathbf{0}$ \\
								Effect size & $-0.2389$ & $-0.0716$ & $-0.0412$ & $-0.0103$ & $-0.0016$ & $0.0102$ & $0.0049$ & $-0.0145$ & $^\dagger\mathbf{-1.1708}$ & $^\dagger\mathbf{-1.1787}$ \\
								\bottomrule
							\end{tabular}
						\end{adjustbox}
				\end{sc}
			\end{sc}
		\end{small}
	\end{center}
	\vskip -0.1in
\end{table*}

\textbf{Statistical analysis.}\footnote{Although the results appear to be multimodal, our non-parametric tests are based on ranking and therefore do not make assumptions regarding the underlying distribution. Our tests are therefore still appropriate.
}
Using the Iman-Davenport omnibus test, we reject the null hypothesis of no difference between the different initialisations for both training speed and generalisation, with a $p$-value equal to $2.2 \times 10^{-16}$ (practically zero). 
This is somewhat unsurprising, since there are clear differences between the initialisations closer to criticality and those at the extremes (E1 and E2). 
Therefore, given that we have rejected $\textrm{H}_0$, we also conduct post-hoc tests. 

Table \ref{tab: h tests} provides pairwise comparisons between the critical initialisation and the alternatives. 
For mean training speed, we find that only the initialisations at the extremes, i.e. close to $\sigma^2_w(\alpha)$ and $\sigma^2_w(\beta)$, give significantly different results. 
These include initialising very close to zero (L4) and very large initialisations (E1 and E2). 
For the initialisations around criticality the differences are not statistically significant. 
For generalisation, it seems that the alternative initialisations are more sensitive to deviations from criticality (only R1 and L3 indicate no statistically significant difference). 
However, given the large scale of our study we are able to detect very fine differences. 
Therefore, even when differences are significant, it is important to consider the sizes of their effects. 

\textbf{Effect sizes.} The purpose of computing effect sizes is to gauge whether statistically significant differences in effects are actually meaningful as measured by their magnitude. 
For a metric $\tau$, we define the effect size for an alternative initialisation $a \in \mathcal{A}$, as $d_a(\tau) = [\mu_a(\tau) - \mu_{\textrm{crit}}(\tau)] / \textrm{sd}_{\textrm{crit}}(\tau)$, where $\textrm{sd}_{\textrm{crit}}(\tau)$ is the standard deviation of $\tau$ for the critical initialisation. 
This definition of effect size for a given quantity is often referred to as Cohen's $d$ \citep{cohen1988}. 
In our context, a value of $d=1$ can be interpreted as a difference in effects equal to one standard deviation away from the mean of criticality. 
Effect sizes are typically considered to be large, i.e. meaningful, for $d \geq 0.8$. The effect sizes for all the alternative initialisations are given in Table \ref{tab: h tests}, where the direction of an effect is indicated by its sign (negative indicating a worse performance when compared to criticality). 
Effect sizes larger than $0.8$ in absolute value are marked with the $\dagger$ symbol. 
As suggested by the plots in Figure \ref{fig: results}, the majority of initialisations around criticality with statistically significant differences in generalisation, as shown in Table \ref{tab: h tests}, have negligible effects sizes. 
The only meaningful effects are again at the extremes. 
Finally, we note that the above findings also hold when just considering standard ReLU networks without dropout (shown in Appendix D).

%% file: conclusion.tex
\section{Conclusion}

At large depth, critical initialisation for neural networks is often considered crucial for success in training and generalisation. 
However, recent work has shown that dropout, a popular regularisation strategy, limits the depth to which networks can be trained. 
Given this depth limit, we explore the question of whether initialising at criticality still matters. Or whether it is possible that alternative, non-critical initialisations (less suited for stable signal propagation at extreme depth) provide any previously undiscovered benefits over critical initialisation.
We conducted a large-scale controlled experiment by training over $12000$ neural networks. 
A systematic statistical analysis of training speed and generalisation performance showed that, for a wide range of alternative initialisations around criticality, there is no statistically significant difference between these initialisations and the critical initialisation.
Our analysis provides strong evidence that, for moderately deep feedforward ReLU networks (as well as those whose depth is constrained by dropout), there is little to be gained by searching for alternative initialisation schemes.

We emphasize the value of the methodology presented in this paper. 
Initialisation aside, the methodology can be followed in a generic way to rigorously test the effects of any design component of interest associated with a particular machine learning algorithm.
Since statistical rigour is often lacking in empirical machine learning research, we hope that this approach might serve as a useful template for more rigorous future investigations.

%% file: suppmat.tex
\appendix
\section*{Appendix}
\addcontentsline{toc}{section}{Appendices}
\renewcommand{\thesubsection}{\Alph{subsection}}

\subsection{Pseudo random design construction}
\subsubsection{Initial / stage-one designs}
First we construct incomplete neural network designs by taking the random Cartesian product of the form:
\begin{align}
    \Omega_* = \mathcal{W} \times \mathcal{B} \times \mathcal{O} \times \mathcal{M} \times \mathcal{L}.
\end{align}

To ensure balanced designs are created (in the sense that the design space contains an equal number of each item in any of the above sets), we do not sample from the sets with a uniform random probability of selecting each element, but rather concatenate random permutations of each set and pair configurations across this.
This process is best illustrated with a small example:

Suppose we only have two hyper-parameter choices: the width of the hidden layers and the learning rate.
We would then want to generate pairs of layer-width-learning-rate samples.
Our method is to:
\begin{enumerate}
    \item concatenate random permutations of each set: \\
    $\mathcal{\widehat{W}} = \left[ \text{permute}\left( \mathcal{W} \right), \dots , \text{permute}\left( \mathcal{W} \right) \right] =\\ \left[ 600, 400, 800, 800, 600, 400, 600, 400 \right]$ (for example) \\
    $\mathcal{\widehat{L}} = \left[ \text{permute}\left( \mathcal{L} \right), \dots , \text{permute}\left( \mathcal{L} \right) \right] =\\ \left[ 10^{-4}, 10^{-3}, 10^{-5}, 10^{-6}, 10^{-3}, 10^{-6}, 10^{-5}, 10^{-4} \right]$ (for example)
    \item sequentially pair these concatenated sets:\\
    $\Omega_*^{(i)} = \left( \mathcal{\widehat{W}}^{(i)}, \mathcal{\widehat{L}}^{(i)} \right) \\
    \therefore \Omega_*^{(0)} = \left( 600, 10^{-4} \right); \Omega_i^{(1)} = \left( 400, 10^{-3} \right); $ etc
\end{enumerate}

We then duplicate these combinations for each dropout rate we wish to test.
Subsequently, we generate the set of viable correlation information preserving depths based on the dropout rate that is present in each configuration.
We use the same setup as above to pair viable depths with incomplete combinations to form complete combinations.

Note that Adam does not support momentum. Thus, when Adam and momentum values were paired, the momentum parameter was ignored when creating the network.

\begin{figure}
	\centering
	\includegraphics[width=0.8\linewidth]{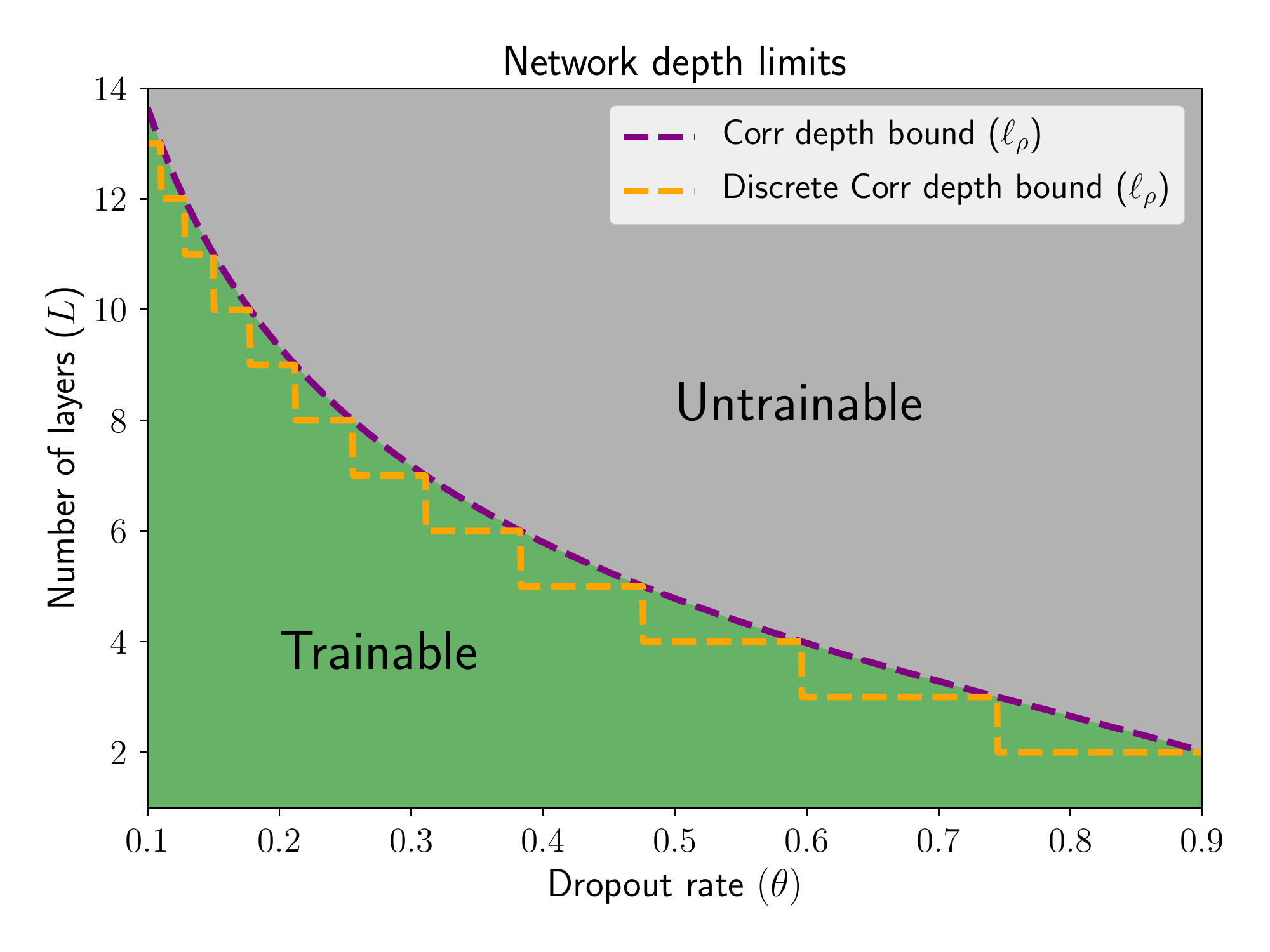}
	\caption{\textit{Discrete trainable depth boundary}. The depth to which networks with dropout are trainable for different dropout rates (dashed purple line). Discrete depths for each dropout rate is shown by the dashed orange line. When creating networks in practice, the discrete bound should be considered.}
	\label{fig: descrete theory}
\end{figure}

\subsubsection{Complete / stage-two designs}
A process identical to the above is followed to match incomplete designs $\Omega_*$ with viable network depths based on the dropout rate of the group.
We construct depth sets $\mathcal{D}_\theta \subset \mathcal{D}$ such that $\mathcal{D}_\theta$ only contains depths to which correlation information can propagate for the given dropout rate $\theta$ --see Figure \ref{fig: descrete theory} for a graphical representation of this and note that, since we are working with networks with discrete numbers of layers, we follow the discretised boundary.
We then concatenate permutations of this set and use this to complete the designs and form our final designs $\Omega$.
Let us illustrate this by continuing the above example for $\theta = 0.7$:

\begin{enumerate}
    \item construct $\mathcal{D}_{0.7} = \{2, 3, 4, 5, 6, 7\}$ using (6) (we omit networks with 0 and 1 hidden layers to ensure some measure of expressiblity)
    \item concatenate random permutations of $\mathcal{D}_{0.7}$: \\
     $\mathcal{\widehat{D}}_{0.7} = \left[ \text{permute}\left( \mathcal{D}_{0.7} \right), \dots , \text{permute}\left( \mathcal{D}_{0.7} \right) \right] =\\ \left[ 6, 3, 7, 2, 5, 4, 2, 3, 6, 5, 4, 7 \right]$ (for example)
     \item sequentially pair this concatenated set with the incomplete designs: \\
     $\Omega^{(i)} = \left( \Omega_*^{(i)}, \mathcal{\widehat{D}}_{0.7}^{(i)}, 0.7 \right) \\
    \therefore \Omega^{(0)} = \left( 600, 10^{-4}, 6, 0.7 \right); \Omega^{(1)} = \left( 400, 10^{-3}, 3, 0.7 \right); $ etc
\end{enumerate}

These four design sets, one for each value in $\mathcal{R}$, are then duplicated $44$ times (once for each combination of initialisation candidate in $\mathcal{A}$ and dataset in $\mathcal{X}$).

\subsection{Method for generating network design dependent initialisations}

Inputs:
\begin{itemize}
    \item $\theta$: dropout rate
    \item $L$: depth of the network
    \item $S$: the number of candidates on either side of critical, within the core group, to be generated
    \item $E$: the number of ``extreme'' candidates (those far greater than critical) to be generated
    \item $\beta$: the largest positive value that can be represented given the floating point precision of the current computer
    \item $\alpha$: the smallest positive value that can be represented given the floating point precision of the current computer
\end{itemize}

Steps:
\begin{enumerate}
    \item calculate $\sigma_{\textrm{critical}}^2 = 2(1 - \theta)$
    \item calculate $\sigma_{w}^2 \left( \beta \right)$ using (10)
    \item generate the set of ``extreme'' samples, $\mathbf{\sigma_{extreme}^2}$:
    \begin{enumerate}
        \item select the first ``extreme'' candidate such that it is within the depth boundary: \\
        $\sigma_{\text{extreme}, E}^2 = 0.9 \sigma_{w}^2 \left( \beta \right)$
        \item recursively calculate the subsequent "extreme" candidates such that they are logarithmically spaced: $\sigma_{\text{extreme}, e}^2 = \frac{1}{2} \sigma_{\text{extreme}, (e + 1)}^2$ for $e \in \left\{1, 2, ..., E - 1\right\}$
    \end{enumerate}
    \item calculate $\sigma_{w}^2 \left( \alpha \right)$ using (9)
    \item generate the set of logarithmically spaced samples less than critical, $\mathbf{\sigma_{left}^2}$: \\
    $\sigma_{\text{left}, s}^2 = \sigma_{\text{critical}}^2 - \frac{0.9}{2^{s - 1}} (\sigma_{\text{critical}}^2 - \sigma_{w}^2 \left( \alpha \right))$ for $s \in \left\{1, 2, ..., S \right\}$
    \item generate the set of samples just greater than critical, $\mathbf{\sigma_{right}^2}$, by reflecting $\mathbf{\sigma_{left}^2}$ about the critical initialisation: \\
    $\sigma_{right, s}^2 = \sigma_{\text{critical}}^2 - (\sigma_{left, s}^2 - \sigma_{\text{critical}}^2) = \sigma_{\text{critical}}^2 + \frac{0.9}{2^{s - 1}} (\sigma_{\text{critical}}^2 - \sigma_{w}^2 \left( \alpha \right))$ for $s \in \left\{1, 2, ..., S \right\}$
    
\end{enumerate}

The set of candidate initialisations is then $\left\{ \mathbf{\sigma_{left}^2}, \sigma_{\text{critical}}^2, \mathbf{\sigma_{right}^2}, \mathbf{\sigma_{extreme}^2} \right\}$.

\subsection{Design and corresponding initialisation examples}
Table \ref{tab: example designs} shows 12 sampled designs and their corresponding initialisations.
These samples are representative of our full set of design samples and give a good idea of typical network parameters.
While there appears to be no difference between core initialisation values across samples with the same dropout rate, this is actually not the case.
Changes are simply typically too small to be seen with only 3 decimal places.
This is due to the rate of change of $\sigma_w^2(\alpha)$ being very low for networks of shallow to moderate depth (roughly 20 hidden layers or less).

\begin{table*}
	\caption{12 example sets of sampled designs and their corresponding initialisations.}
	\label{tab: example designs}
	\begin{center}
		\begin{scriptsize}
			\begin{sc}
				-- Example designs --
				\vspace{0.2cm}
			    \begin{adjustbox}{width=\linewidth,center}
				    \begin{tabular}{ccccccccc}
                    \toprule
                    design index & dataset & depth & width & rate & batch size & optimiser & momentum & learning rate \\
                    \toprule
                    $0$ & CIFAR-10 & $12$ & $800$ & $0.0$ & $256$ & Adam & $0.0$ & $10^{-5}$ \\
                    $1$ & FashionMNIST & $15$ & $800$ & $0.0$ & $64$ & SGD & $0.5$ & $10^{-6}$ \\
                    $2$ & MNIST & $4$ & $400$ & $0.0$ & $256$ & RMSprop & $0.0$ & $10^{-5}$ \\
                    $3$ & CIFAR-10 & $2$ & $400$ & $0.5$ & $256$ & RMSprop & $0.0$ & $10^{-5}$ \\
                    $4$ & CIFAR-100 & $3$ & $800$ & $0.5$ & $64$ & SGD & $0.5$ & $10^{-6}$ \\
                    $5$ & FashionMNIST & $4$ & $800$ & $0.5$ & $256$ & Adam & $0.0$ & $10^{-5}$ \\
                    $6$ & CIFAR-10 & $7$ & $600$ & $0.3$ & $256$ & Adam & $0.0$ & $10^{-3}$ \\
                    $7$ & CIFAR-100 & $4$ & $600$ & $0.3$ & $64$ & SGD & $0.5$ & $10^{-6}$ \\
                    $8$ & FashionMNIST & $5$ & $400$ & $0.3$ & $256$ & RMSprop & $0.0$ & $10^{-5}$ \\
                    $9$ & CIFAR-10 & $3$ & $600$ & $0.1$ & $32$ & SGD & $0.5$ & $10^{-4}$ \\
                    $10$ & MNIST & $8$ & $600$ & $0.1$ & $32$ & SGD & $0.5$ & $10^{-4}$ \\
                    $11$ & CIFAR-10 & $4$ & $600$ & $0.1$ & $128$ & Adam & $0.0$ & $10^{-3}$ \\
                    \bottomrule
                    \end{tabular}
				\end{adjustbox}
				
				\begin{sc}
					-- Initialisations --
                    \begin{adjustbox}{width=\linewidth,center}
                        \begin{tabular}{cccccccccccc}
                        \toprule
                        design index & L4 & L3 & L2 & L1 & C & R1 & R2 & R3 & R4 & E1 & E2 \\
                        \toprule
                        $0$  & $0.201$ & $1.101$ & $1.550$ & $1.775$ & $2.000$ & $2.225$ & $2.450$ & $2.899$ & $3.799$ & $1.464 \times 10^{3}$ &  $2.926 \times 10^{3}$ \\
                        $1$  & $0.205$ & $1.103$ & $1.551$ & $1.776$ & $2.000$ & $2.224$ & $2.449$ & $2.897$ & $3.795$ & $3.346 \times 10^{2}$ &  $6.671 \times 10^{2}$ \\
                        $2$  & $0.200$ & $1.100$ & $1.550$ & $1.775$ & $2.000$ & $2.225$ & $2.450$ & $2.900$ & $3.800$ & $3.865 \times 10^{9}$ &  $7.731 \times 10^{9}$ \\
                        $3$  & $0.100$ & $0.550$ & $0.775$ & $0.887$ & $1.000$ & $1.113$ & $1.225$ & $1.450$ & $1.900$ & $8.301 \times 10^{18}$ & $1.660 \times 10^{19}$ \\
                        $4$  & $0.100$ & $0.550$ & $0.775$ & $0.888$ & $1.000$ & $1.112$ & $1.225$ & $1.450$ & $1.900$ & $3.142 \times 10^{12}$ & $6.283 \times 10^{12}$ \\
                        $5$  & $0.100$ & $0.550$ & $0.775$ & $0.888$ & $1.000$ & $1.112$ & $1.225$ & $1.450$ & $1.900$ & $1.933 \times 10^{9}$ &  $3.865 \times 10^{9}$ \\
                        $6$  & $0.140$ & $0.770$ & $1.085$ & $1.243$ & $1.400$ & $1.557$ & $1.715$ & $2.030$ & $2.660$ & $2.013 \times 10^{5}$ &  $4.026 \times 10^{5}$ \\
                        $7$  & $0.140$ & $0.770$ & $1.085$ & $1.243$ & $1.400$ & $1.557$ & $1.715$ & $2.030$ & $2.660$ & $2.706 \times 10^{9}$ &  $5.412 \times 10^{9}$ \\
                        $8$  & $0.140$ & $0.770$ & $1.085$ & $1.243$ & $1.400$ & $1.557$ & $1.7154$ & $2.030$ & $2.660$ & $3.204 \times 10^{7}$ &  $6.408 \times 10^{7}$ \\
                        $9$  & $0.180$ & $0.990$ & $1.395$ & $1.598$ & $1.800$ & $2.002$ & $2.205$ & $2.610$ & $3.4204$ & $5.655 \times 10^{12}$ & $1.131 \times 10^{13}$ \\
                        $10$ & $0.180$ & $0.990$ & $1.395$ & $1.598$ & $1.800$ & $2.002$ & $2.205$ & $2.610$ & $3.420$ & $5.309 \times 10^{4}$ & $1.062 \times 10^{5}$ \\
                        $11$ & $0.180$ & $0.990$ & $1.395$ & $1.598$ & $1.800$ & $2.002$ & $2.205$ & $2.610$ & $3.420$ & $3.479 \times 10^{9}$ & $6.958 \times 10^{9}$ \\
                        \bottomrule
                        \end{tabular}
                    \end{adjustbox}
				\end{sc}
			\end{sc}
		\end{scriptsize}
	\end{center}
	\vskip -0.1in
\end{table*}

\subsection{Additional results}

In this section we provide additional statistical analyses per dropout rate as well as with zero dropout. 
These results are given in Tables \ref{tab: no drop}, \ref{tab: drop 0.1}, \ref{tab: drop 0.3} and \ref{tab: drop 0.5}.

\begin{table*}
	\caption{\textbf{No dropout -- $\theta = 0$}: Post-hoc tests for training speed and generalisation for no dropout ($\theta = 0$). The symbols $^*$ indicate a significant $p$-value and $^\dagger$ a large effect size.}
	\label{tab: no drop}
	\begin{center}
		\begin{small}
			\begin{sc}
				Training speed -- $\textrm{H}_0: \mu_{\textrm{crit}}(\tau_s) = \mu_a(\tau_s), a \in \{L_i, R_i, E_1, E_2\}^4_{i=1}$
				\vspace{0.2cm}
			    \begin{adjustbox}{width=\linewidth,center}
				    \begin{tabular}{ccccccccccc}
					    \toprule
					    & L4 & L3 & L2 & L1 & R1 & R2 & R3 & R4 & E1 & E2 \\
						\toprule
						Adjusted $p$-value & $^*\mathbf{0}$ & $\mathbf{1.99\times10^{-11}}$  & $\mathbf{3.07\times10^{-5}}$ & $0.0864$ & $0.0529$ & $\mathbf{9.68\times10^{-5}}$ & $\mathbf{8.89\times10^{-6}}$ & $\mathbf{4.71\times10^{-7}}$ & $^*\mathbf{0}$ & $^*\mathbf{0}$ \\
						Effect size & $-0.6024$ & $-0.1854$ & $-0.1124$ & $-0.0434$ & $0.0382$ & $0.0716$ & $0.0774$ & $0.0602$ & $^\dagger\mathbf{-1.1085}$ & $^\dagger\mathbf{-1.1009}$ \\
					    \bottomrule
					\end{tabular}
				\end{adjustbox}
				
				\begin{sc}
					Generalisation -- $\textrm{H}_0: \mu_{\textrm{crit}}(\tau_g) = \mu_a(\tau_g), a \in \{L_i, R_i, E_1, E_2\}^4_{i=1}$ \\
						\begin{adjustbox}{width=\linewidth,center}
							\begin{tabular}{cccccccccccc}
								\toprule
								& L4 & L3 & L2 & L1 & R1 & R2 & R3 & R4 & E1 & E2 \\
								\toprule
								Adjusted $p$-value & $^*\mathbf{0}$ & $^*\mathbf{0.0033}$ & $0.8549$ & $0.7927$  &  $0.1846$ & $^*\mathbf{0.0018}$ & $^*\mathbf{7.20\times10^{-6}}$ & $^*\mathbf{3.49\times10^{-14}}$ & $^*\mathbf{0}$ & $\mathbf{0}$ \\
								Effect size & $-0.5224$ & $-0.1592$ & $-0.0975$ & $-0.0325$ & $-0.0027$ &  $0.0077$ & $-0.0032$ & $-0.0374$ & $^\dagger \mathbf{-1.0008}$ & $^\dagger\mathbf{-1.0191}$ \\
								\bottomrule
							\end{tabular}
						\end{adjustbox}
				\end{sc}
			\end{sc}
		\end{small}
	\end{center}
	\vskip -0.1in
\end{table*}

\begin{table*}
	\caption{\textbf{Dropout -- $\theta = 0.1$}: Post-hoc tests for training speed and generalisation for dropout with rate $\theta = 0.5$. The symbols $^*$ indicate a significant $p$-value and $^\dagger$ a large effect size.}
	\label{tab: drop 0.1}
	\begin{center}
		\begin{small}
			\begin{sc}
				Training speed -- $\textrm{H}_0: \mu_{\textrm{crit}}(\tau_s) = \mu_a(\tau_s), a \in \{L_i, R_i, E_1, E_2\}^4_{i=1}$
				\vspace{0.2cm}
			    \begin{adjustbox}{width=\linewidth,center}
				    \begin{tabular}{ccccccccccc}
					    \toprule
					    & L4 & L3 & L2 & L1 & R1 & R2 & R3 & R4 & E1 & E2 \\
						\toprule
						Adjusted $p$-value & $^*\mathbf{0}$ & $^*\mathbf{0.0203}$ & $0.9219$ & $0.6552$ & $0.9395$ & $0.2258$ & $^*\mathbf{0.0021}$ & $^*\mathbf{2.59\times10^{-8}}$ &  $^*\mathbf{0}$ & $^*\mathbf{0}$ \\
						Effect size & $-0.2578$ & $-0.0723$ & $-0.0189$ &  $0.0015$ & $0.0311$ &  $0.0261$ &  $0.0202$ & $-0.0009$ & $^\dagger\mathbf{-1.1328}$ & $^\dagger\mathbf{-1.1209}$ \\
					    \bottomrule
					\end{tabular}
				\end{adjustbox}
				
				\begin{sc}
					Generalisation -- $\textrm{H}_0: \mu_{\textrm{crit}}(\tau_g) = \mu_a(\tau_g), a \in \{L_i, R_i, E_1, E_2\}^4_{i=1}$ \\
						\begin{adjustbox}{width=\linewidth,center}
							\begin{tabular}{cccccccccccc}
								\toprule
								& L4 & L3 & L2 & L1 & R1 & R2 & R3 & R4 & E1 & E2 \\
								\toprule
								Adjusted $p$-value & $^*\mathbf{0}$ & $0.4475$ & $0.8743$ & $0.7033$ & $0.7874$ & $0.5100$ & $^*\mathbf{0.0193}$ & $^*\mathbf{1.41\times10^{-6}}$ & $^*\mathbf{0}$ & $^*\mathbf{0}$ \\
								Effect size & $-0.3091$ & $-0.0976$ & $-0.0650$ & $-0.0237$ & $0.0139$ &  $0.0236$ &  $0.0221$ &  $0.0167$ & $^\dagger\mathbf{-1.1503}$ & $^\dagger\mathbf{-1.1572}$\\
								\bottomrule
							\end{tabular}
						\end{adjustbox}
				\end{sc}
			\end{sc}
		\end{small}
	\end{center}
	\vskip -0.1in
\end{table*}

\begin{table*}
	\caption{\textbf{Dropout -- $\theta = 0.3$}: Post-hoc tests for training speed and generalisation for dropout with rate $\theta = 0$. The symbols $^*$ indicate a significant $p$-value and $^\dagger$ a large effect size.}
	\label{tab: drop 0.3}
	\begin{center}
		\begin{small}
			\begin{sc}
				Training speed -- $\textrm{H}_0: \mu_{\textrm{crit}}(\tau_s) = \mu_a(\tau_s), a \in \{L_i, R_i, E_1, E_2\}^4_{i=1}$
				\vspace{0.2cm}
			    \begin{adjustbox}{width=\linewidth,center}
				    \begin{tabular}{ccccccccccc}
					    \toprule
					    & L4 & L3 & L2 & L1 & R1 & R2 & R3 & R4 & E1 & E2 \\
						\toprule
						Adjusted $p$-value & $^*\mathbf{0.0051}$ & $0.0727$ & $0.0727$ & $0.2097$ & $0.1885$ & $^*\mathbf{0.0090}$ & $^*\mathbf{1.49\times10^{-6}}$ & $^*\mathbf{4.88\times10^{-14}}$ & $^*\mathbf{0}$ & $^*\mathbf{0}$ \\
						Effect size & $-0.1348$ & $-0.0281$ & $-0.0020$ & $-0.0009$ & $-0.0066$ & $-0.0047$ & $-0.0124$ & $-0.0336$ & $^\dagger\mathbf{-1.1267}$ & $^\dagger\mathbf{-1.1304}$\\
					    \bottomrule
					\end{tabular}
				\end{adjustbox}
				
				\begin{sc}
					Generalisation -- $\textrm{H}_0: \mu_{\textrm{crit}}(\tau_g) = \mu_a(\tau_g), a \in \{L_i, R_i, E_1, E_2\}^4_{i=1}$ \\
						\begin{adjustbox}{width=\linewidth,center}
							\begin{tabular}{cccccccccccc}
								\toprule
								& L4 & L3 & L2 & L1 & R1 & R2 & R3 & R4 & E1 & E2 \\
								\toprule
								Adjusted $p$-value & $^*\mathbf{0.0409}$ & $^*\mathbf{0.0018}$ & $^*\mathbf{0.0075}$ & $^*\mathbf{0.0409}$ & $0.3975$ & $^*\mathbf{0.0296}$ & $^*\mathbf{1.03\times10^{-5}}$ & $^*\mathbf{6.88\times10^{-14}}$ & $^*\mathbf{0}$ & $^*\mathbf{0}$ \\
								Effect size & $-0.1144$ & $-0.0416$ & $-0.0074$ &  $0.0039$ &  $0.0035$ &  $0.0093$ &  $0.0128$ & $-0.0113$ & $^\dagger\mathbf{-1.2549}$ & $^\dagger\mathbf{-1.2575}$\\
								\bottomrule
							\end{tabular}
						\end{adjustbox}
				\end{sc}
			\end{sc}
		\end{small}
	\end{center}
	\vskip -0.1in
\end{table*}

\begin{table*}
	\caption{\textbf{Dropout -- $\theta = 0.5$}: Post-hoc tests for training speed and generalisation for dropout with rate $\theta = 0$. The symbols $^*$ indicate a significant $p$-value and $^\dagger$ a large effect size.}
	\label{tab: drop 0.5}
	\begin{center}
		\begin{small}
			\begin{sc}
				Training speed -- $\textrm{H}_0: \mu_{\textrm{crit}}(\tau_s) = \mu_a(\tau_s), a \in \{L_i, R_i, E_1, E_2\}^4_{i=1}$
				\vspace{0.2cm}
			    \begin{adjustbox}{width=\linewidth,center}
				    \begin{tabular}{ccccccccccc}
					    \toprule
					    & L4 & L3 & L2 & L1 & R1 & R2 & R3 & R4 & E1 & E2 \\
						\toprule
						Adjusted $p$-value &  $0.1170$ & $^*\mathbf{7.46\times10^{-5}}$ & $^*\mathbf{0.0075}$ & $0.0843$ & $0.1188$ & $^*\mathbf{0.0002}$ & $^*\mathbf{1.63\times10^{-8}}$ & $^*\mathbf{2.66\times10^{-14}}$ & $^*\mathbf{0}$ & $^*\mathbf{0}$ \\
						Effect size & $-0.0491$ &  $0.01421$ &  $0.0105$ &  $0.0119$ & $-0.0227$ & $-0.0073$ & $-0.0210$ & $-0.0404$ & $^\dagger\mathbf{-1.1158}$ & $
						^\dagger\mathbf{-1.1171}$ \\
					    \bottomrule
					\end{tabular}
				\end{adjustbox}
				
				\begin{sc}
				 Generalisation -- $\textrm{H}_0: \mu_{\textrm{crit}}(\tau_g) = \mu_a(\tau_g), a \in \{L_i, R_i, E_1, E_2\}^4_{i=1}$ \\
						\begin{adjustbox}{width=\linewidth,center}
							\begin{tabular}{cccccccccccc}
								\toprule
								& L4 & L3 & L2 & L1 & R1 & R2 & R3 & R4 & E1 & E2 \\
								\toprule
								Adjusted $p$-value & $^*\mathbf{0.0009}$ & $^*\mathbf{6.18\times10^{-6}}$ & $^*\mathbf{0.0016}$ & $0.0591$ & $0.2637$ & $^*\mathbf{0.0010}$ & $^*\mathbf{1.03\times10^{-5}}$ & $^*\mathbf{5.08\times10^{-12}}$ & $^*\mathbf{0}$ & $^*\mathbf{0}$ \\
								Effect size & $-3.37\times10^{-2}$ &  $4.44\times10^{-3}$ & $1.01\times10^{-5}$ & $8.92\times10^{-3}$ & $-2.04\times10^{-2}$ & $4.63\times10^{-4}$ & $-1.18\times10^{-2}$ & $-2.64e-02$ & $^\dagger\mathbf{-1.2796}$ & $^\dagger\mathbf{-1.2843}$ \\
								\bottomrule
							\end{tabular}
						\end{adjustbox}
				\end{sc}
			\end{sc}
		\end{small}
	\end{center}
	\vskip -0.1in
\end{table*}

\subsection{On the validity of the RCT approach}

We performed two auxiliary studies to ensure the effectiveness of the RCT setup.

Firstly, we wanted to ensure the methodology was set up correctly and could identify known performance differences.
To achieve this, we created a smaller scale scenario very similar to the study described in the main text but instead using initialisation as ``intervention'', we use the activation function.
Networks with non-linear activation functions are more expressive and should be able to outperform their linear counterparts.
Furthermore, the ReLU activation function does not suffer from saturation or vanishing gradients and typically outperforms the sigmoid when using random Gaussian initialisation.
These are well established results, frequently demonstrated in the literature. 
Therefore, we decided to construct an RCT where the interventions are the following activations: linear, sigmoid, and ReLU.
Each network was initialised critically and trained on MNIST for $1955$ iterations using a batch size of $128$.
This RCT consisted of $1761$ trained networks.

\begin{figure}
	\centering
	\includegraphics[width=0.9\linewidth]{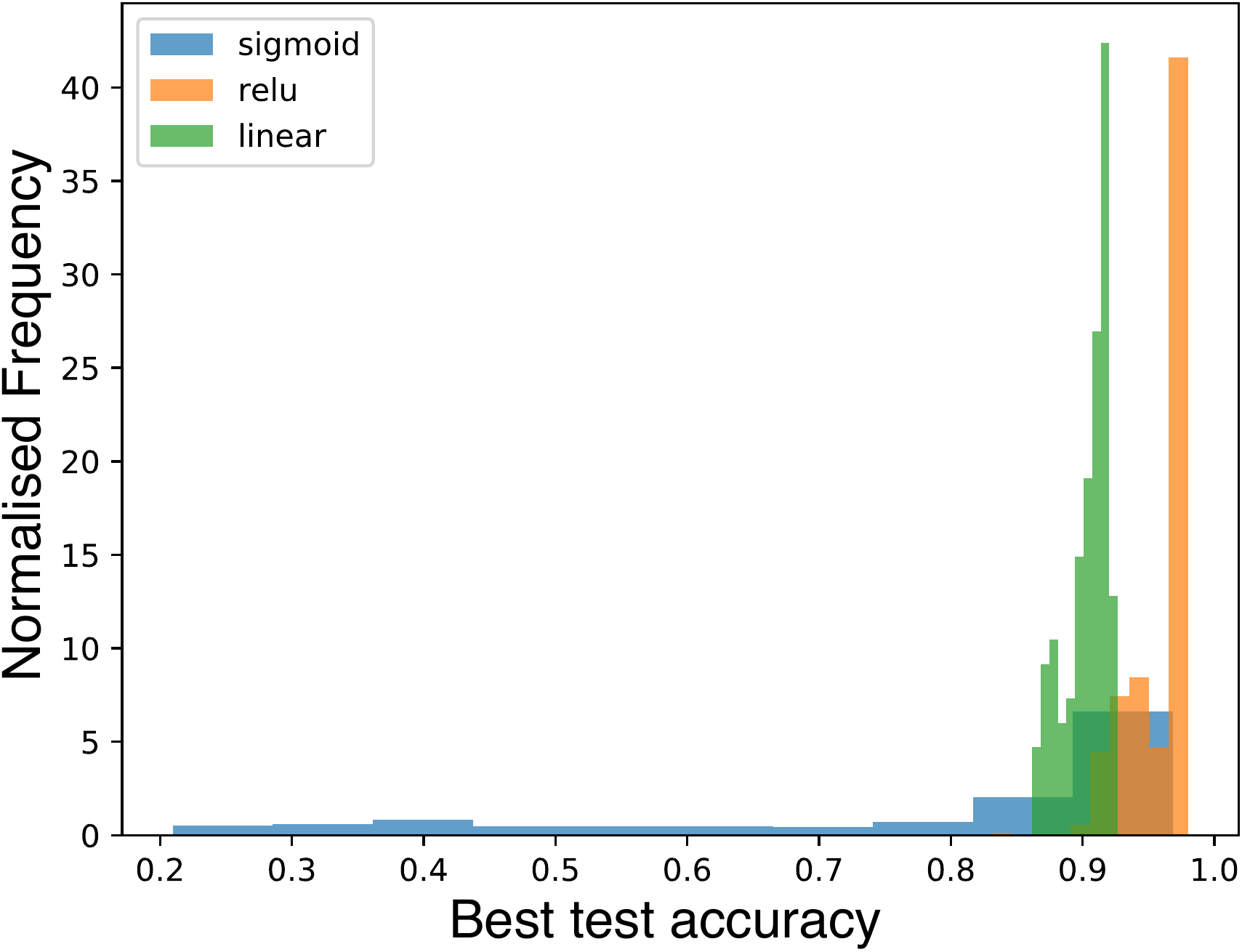}
	\caption{\textit{Best test accuracy distribution per activation function}. The test RCT confirms the validity of this setup as it clearly confirms well known results: (1) ReLU networks typically outperform linear and sigmoidal networks, (2) the best performing sigmoidal networks outperform linear networks and perform comparatively to ReLU networks, but are typically more difficult to train due to vanishing gradients.}
	\label{fig: activation study}
\end{figure}

The results of the above experiment are exactly as expected and are given in Figure \ref{fig: activation study}.
ReLU networks performed best overall.
The performance of linear networks were capped significantly below that of ReLU networks.
Finally, the sigmoidal networks were able to perform better than linear networks and as well as ReLU networks in the best case.
However, the distribution over performance for the sigmoid exhibits a long tail towards low accuracy due to vanishing gradient issues, causing network training to stall.
Furthermore, pairwise comparisons with post-hoc tests between ReLU and the other activations yield $p$-values that are practically zero, indicating significantly different mean performances with meaningful effects sizes ($-2.65$ for linear and $-8.42$ for sigmoid).

After confirming the validity of our RCT setup, we aimed to control for more sophisticated training procedures, such as learning rate decay. 
Learning rate decay is expected to have a complex temporal interaction with other design components during training and poses a challenge as a potential confounder for the RCT to control.
We theorise that adding any mechanism that generally improves performance, such as learning rate decay, should have an average effect (taken over sampled designs) that is roughly equal across groups.

\begin{figure*}
	\centering
	\begin{subfigure}
	  \centering
	  \includegraphics[width=0.4\linewidth]{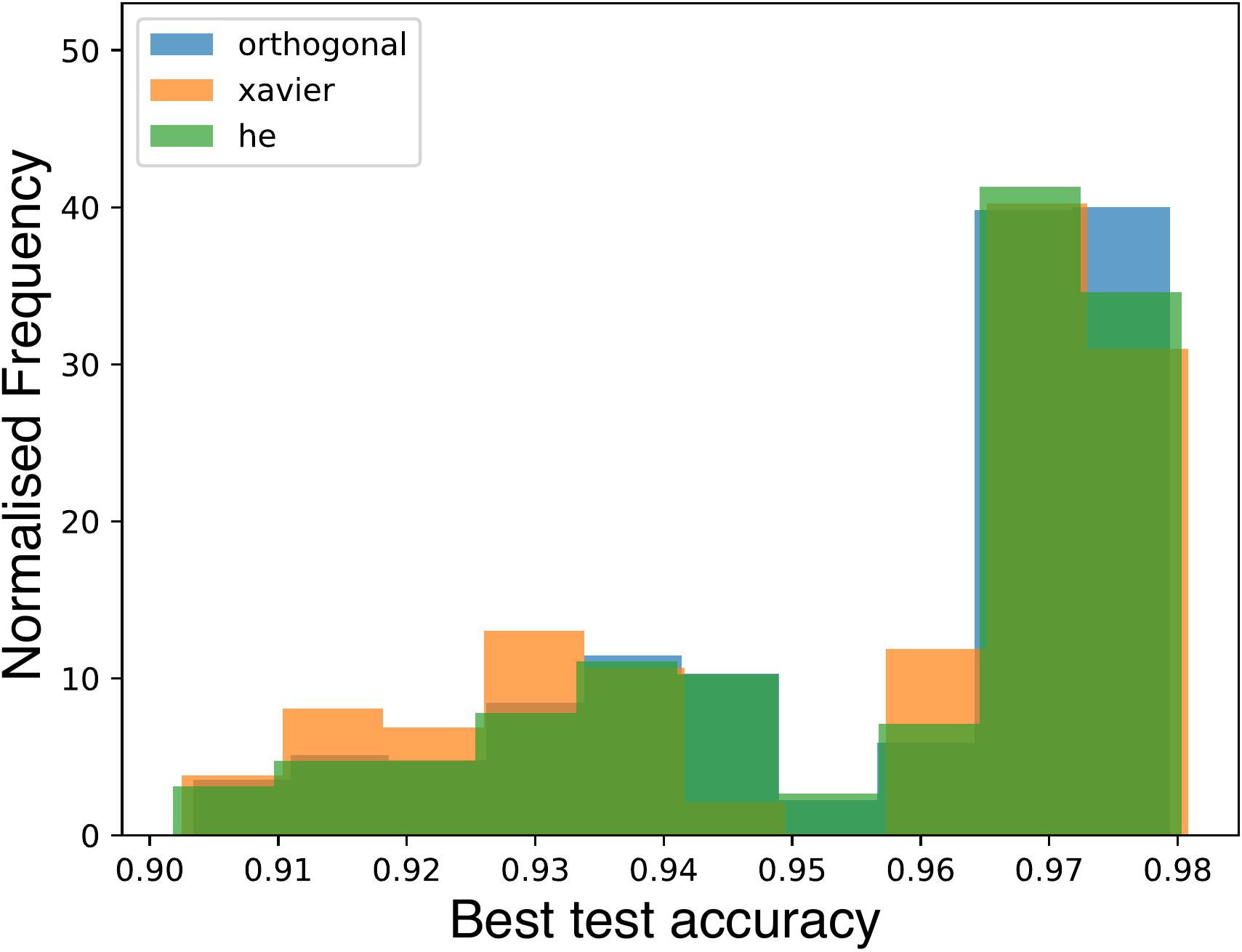}
	  \label{fig:sub1}
	\end{subfigure}%
	\begin{subfigure}
	  \centering
	  \includegraphics[width=0.4\linewidth]{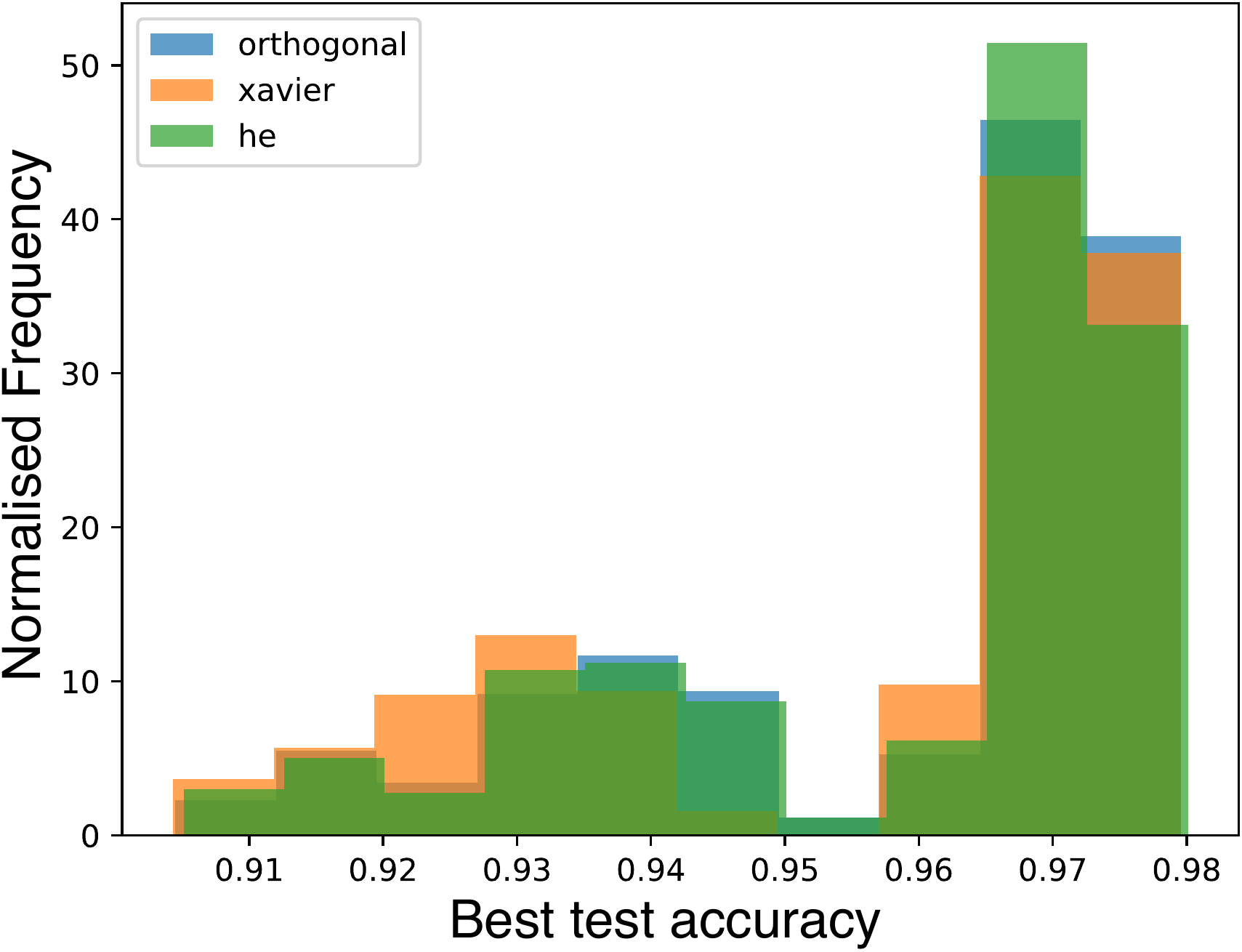}
	  \label{fig:sub2}
	\end{subfigure}
		\caption{\textit{Comparison of RCTs with and without learning rate decay}. \textbf{Left}: Without learning rate decay. \textbf{Right}: With learning rate decay. Best test accuracy distributions per initialisation are shown. It is clear that although learning rate decay improves the overall performance of all groups, the relative performance between groups remains the same.}
	\label{fig: initialisation study}
\end{figure*}

To test this, we construct two RCTs nearly identical to the above.
All networks make use of the ReLU activation function and the interventions in this case are initialisation schemes: He \citep{he2015delving}, Xavier \citep{glorot2010understanding} and Orthogonal \citep{saxe2013exact}.
One of these RCTs makes use of learning rate decay and the other does not.
In this way, we can compare the statistical findings of each and confirm whether they agree. 
If the test results do agree, it means that the RCT has successfully controlled for the confounding effects of learning rate in both cases, i.e. with and without decay.
Figure \ref{fig: initialisation study} shows best test accuracy distributions without and with learning rate decay.
It is clear that although learning rate decay may improve the overall performance of all groups, the relative performance differences between groups remain roughly the same.
Table \ref{tab: learning rate decay} gives the test results for each RCT. 
The conclusions closely match between the two trials.
Therefore, we conclude that the RCT as described and performed in the main text provides a very general approach to isolating the effects of a particular intervention.

A final possible objection to this setup is that samples might not always be independent due to correlations between selected hyperparameters. 
This could be the case when sampling across a coarse grid for a single hyperparameter. 
However, in our setup, we randomly sample over multiple grids of hyperparameters for each design (``participant in our study'').
Thus for correlations between \textit{designs} to persist, they must do so simultaneously across multiple \textit{hyperparameters} (dimensions of the design space) to influence the results. 
Given the high-dimensionality of the design space, we feel it safe to treat each design as an independent sample from the population.

\begin{table}
	\caption{\textbf{Statistical tests using learning rates with and without decay}: Post-hoc tests for generalisation using and not using learning rate decay. Comparisons are between He and Xavier and He and orthogonal initialisation with the null hypothesis $H_0$ of no difference. The relative differences as detected by the tests remain the same between the two approaches, thus the RCT has successfully isolated only the effects of the initialisation.}
	\label{tab: learning rate decay}
	\begin{center}
		\begin{small}
				\begin{sc}
				 Generalisation \\
						\begin{adjustbox}{width=\linewidth,center}
							\begin{tabular}{ccc}
								\toprule
								With decay & Orthogonal & Xavier \\
								\toprule
								Adjusted $p$-value & $0.1972$ & $ ^*\mathbf{0.0904}$ \\
								Effect size & $0.0124$ &  $-0.1176$ \\ 
								\toprule
								Without decay & Orthogonal & Xavier \\
								\toprule
								Adjusted $p$-value & $0.9183$ & $^*\mathbf{0.0067}$ \\
								Effect size & $0.0156$ &  $-0.1130$ 
							\end{tabular}
						\end{adjustbox}
				\end{sc}
		\end{small}
	\end{center}
	\vskip -0.1in
\end{table}
